# Automating Coral Reef Fish Family Identification on Video Transects Using a YOLOv8-Based Deep Learning Pipeline

Jules Gerard
*Marine Biology – Ecology, Evolution & Genetics (bDIV)*
*Vrije Universiteit Brussel (VUB)*
Brussels, Belgium
0009-0007-7989-6493

Leandro Di Bella
*Electronics and Informatics*
*Vrije Universiteit Brussel (VUB)*
Brussels, Belgium
0009-0000-1731-7205

Filip Huyghe
*Marine Biology – Ecology, Evolution & Genetics (bDIV)*
*Vrije Universiteit Brussel (VUB)*
Brussels, Belgium
0009-0000-1731-7205

Marc Kochzius
*Marine Biology – Ecology, Evolution & Genetics (bDIV)*
*Vrije Universiteit Brussel (VUB)*
Brussels, Belgium
0000-0001-8953-4719

*Abstract*—Coral reef monitoring in the Western Indian Ocean is limited by the labor demands of underwater visual censuses. This work evaluates a YOLOv8-based deep learning pipeline for automating family-level fish identification from video transects collected in Kenya and Tanzania. A curated dataset of 24 families was tested under different configurations, providing the first region-specific benchmark for automated reef fish monitoring in the Western Indian Ocean. The best model achieved mAP@0.5 of 0.52, with high accuracy for abundant families but weaker detection of rare or complex taxa. Results demonstrate the potential of deep learning as a scalable complement to traditional monitoring methods.

*Keywords—Coral reef ecology, Fish community monitoring, Western Indian Ocean, Deep Learning, YOLOv8, Underwater imagery.*

## I. Introduction

Coral reefs support critical ecosystem services, including food provision, coastal protection and biodiversity maintenance [1], yet their monitoring remains constrained by the labor intensity and subjectivity of underwater visual censuses (UVCs) [2]. This challenge is particularly acute along the Western Indian Ocean (WIO), the world's second most diverse coral reef region with over 2,400 reef fish species, where data scarcity hampers effective conservation management despite the ecological and socio-economic importance of reef-associated fisheries [3]. Machine learning (ML) and convolutional neural networks (CNNs) offer a promising pathway to scale monitoring capacity [4], but their effectiveness is contingent upon region-specific dataset quality and ecological relevance. Moreover, annotated underwater datasets from the WIO remain rare, making carefully curated, region-specific image collections essential for robust model training and ecological applicability [5].

This work develops and evaluates a YOLOv8 object detection framework to automate coral reef fish identification in East Africa, aiming to improve technical performance and ecological interpretability. The study was guided by three questions: (1) How do dataset-specific factors shape the performance and generalisability of ML-based reef fish detection? (2) How can transfer learning be leveraged to fine-tune YOLOv8 for family-level fish identification in East Africa? (3) What detection performance can be achieved under real-world underwater conditions compared to traditional UVC methods? To address these, the objectives were to: (i) compile and annotate a high-quality family-level dataset from underwater videos collected in Kenya and Tanzania; (ii) fine-tune YOLOv8 using transfer learning and multiple training strategies; and (iii) quantify detection performance with standard computer vision metrics, linking outcomes to monitoring priorities.

## II. Methods

Field data were collected during two campaigns: in 2023 along the Kenyan coast and around Unguja Island (Zanzibar), and in 2025 across Tanzanian mainland reefs and offshore islands, including Chumbe and Tumbatu, spanning a gradient of reef types. At each site, 50 m transects were filmed using GoPro Hero 11 cameras in 5.3K resolution under standardized protocols (total, 81 videos) [6]. Extracted frames (every 3 seconds) were manually annotated in Roboflow at the family level, with selected subfamilies treated separately when morphologically distinctive [7; 8]. Strict bounding-box rules required individuals ≥100 pixels, with ≥50% body visible and clear diagnostic traits. All annotations were conducted and quality-checked, producing 844 images and 5,783 bounding boxes across 24 families (Fig.1), though class imbalance was strong (Pomacentridae 47%). This constitutes one of the first curated, family-level annotated fish datasets developed for the Western Indian Ocean. Several dataset configurations were tested: a full dataset (24 families, Config A), a reduced dataset restricted to the 10 most abundant families (Config B), and a filtered dataset excluding small bounding boxes below 500 px$^2$ (Config C). Models were trained using transfer learning from COCO-pretrained weights, with subsequent hyperparameter tuning under three strategies (default, COCO-derived, and scratch-tuned). Evaluation employed precision, recall, mAP@0.5, mAP@0.5:0.95, and F1-optimised thresholds to balance detection trade-offs.

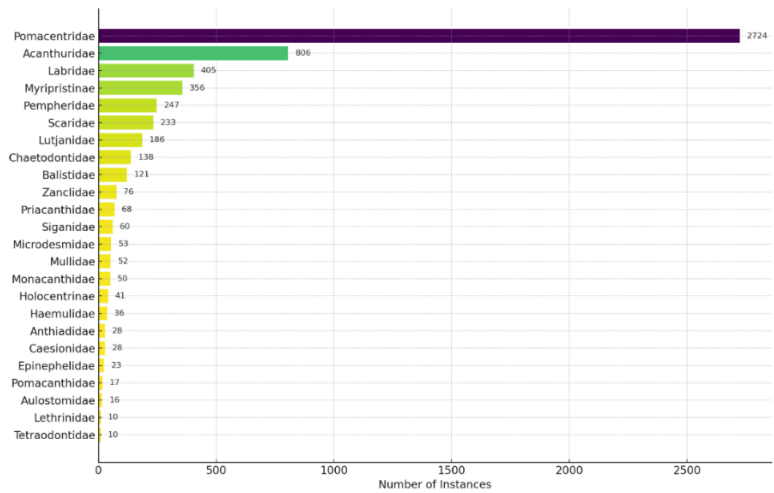

*Figure 1 :Number of annotated instances per fish family in the full dataset (n = 5,783).*

## III. RESULTS

Table 1 summarises the performance of all model configurations. The baseline (Model A, 24 families) achieved moderate precision (0.533) but low recall (0.373) and limited strict accuracy (mAP@0.5:0.95 = 0.250). Restricting training to the 10 most abundant families (Model B) increased recall to 0.460 and mAP@0.5 to 0.465. Refining the dataset further with a 500 px² filter (Model C-DEF) produced the most balanced configuration, raising recall to 0.477 and mAP@0.5 to 0.520. Hyperparameter tuning introduced trade-offs: the scratch-tuned model maximised precision (0.703) but reduced recall, whereas the COCO-tuned model yielded intermediate performance. Applying F1-optimised thresholds further improved detection, with the scratch-tuned variant achieving the highest strict accuracy (mAP@0.5:0.95 = 0.404). At the family level, Myripristinae, Pomacentridae, and Zanclidae consistently achieved high detection scores, whereas Labridae and Lutjanidae remained weak across all models.

TABLE I. MODEL PERFORMANCE METRICS UNDER DIFFERENT CONFIGURATIONS

| Model | Dataset | Performance Metrics | | | |
|---|---|---|---|---|---|
| | | *Precision* | *Recall* | *mAP@0.5* | *mAP@0.5:0.95* |
| A (Full) | 24 families | 0.533 | 0.373 | 0.374 | 0.250 |
| B (Top 10) | 10 families | 0.530 | 0.460 | 0.465 | 0.280 |
| C-DEF | Top 10, ≥500 px² | 0.631 | 0.477 | 0.520 | 0.328 |
| Scratch-tuned | Top 10, ≥500 px² | 0.703 | 0.401 | 0.490 | 0.320 |
| COCO-tuned | Top 10, ≥500 px² | 0.627 | 0.465 | 0.493 | 0.318 |

## IV. CONCLUSION

Results highlight both the feasibility and limitations of deep learning for reef monitoring in the WIO. Detection was reliable for abundant, morphologically distinctive families but limited for rare or visually complex taxa, underscoring the influence of dataset imbalance and morphological similarity [5]. Trade-offs between precision and recall mirrored ecological priorities: high precision is critical for abundance estimation, whereas high recall is essential for detecting rare or endangered taxa [9]. Error analyses revealed systematic challenges from occlusion, small body size and coral background complexity [10; 4]. Importantly, the dataset's restricted size and lack of independent benchmarks (e.g., UVC or eDNA) constrained absolute validation, though the study establishes a reproducible baseline for regional ML applications.

This work demonstrates that YOLOv8 can support consistent presence–absence detection and functional group monitoring for dominant reef fish families in East Africa, thereby offering a scalable complement to traditional UVCs. However, current configurations are not yet suitable for biomass estimation or robust assessments of rare taxa. Future progress hinges on expanding dataset size, improving taxonomic balance and refining filming protocols to minimize occlusion and enhance detection of small-bodied species [11]. Methodological refinements such as targeted data augmentation, tiling approaches and k-fold cross-validation could further strengthen robustness. Integrating object tracking offers a pathway toward abundance estimates, while publishing open datasets and trained weights would enhance reproducibility and enable regional refinement taxa [12]. Collectively, these steps position deep learning as a viable, region-specific tool to enhance coral reef monitoring in the WIO, advancing both ecological understanding and management capacity in a region where monitoring resources are scarce but conservation needs are acute.

ACKNOWLEDGMENT

This research was conducted as part of the Master of Science in Marine and Lacustrine Science and Management (Oceans & Lakes) program in Belgium. This study was funded by the SAVE-FISH project at Vrije Universiteit Brussel (VUB).